\documentclass[letterpaper]{article} 
\usepackage{aaai25}  
\usepackage{times}  
\usepackage{helvet}  
\usepackage{courier}  
\usepackage[hyphens]{url}  
\usepackage{graphicx} 
\usepackage{natbib}  
\usepackage{caption} 
\usepackage{amsmath}
\usepackage{algorithm}
\usepackage{algorithmic}

\usepackage{multirow}
\usepackage{newfloat}
\usepackage{listings}
\usepackage[dvipsnames]{xcolor}
\DeclareCaptionStyle{ruled}{labelfont=normalfont,labelsep=colon,strut=off} 
\floatstyle{ruled}
\newfloat{listing}{tb}{lst}{}
\floatname{listing}{Listing}
\title{Fast Think-on-Graph: Wider, Deeper and Faster \\ Reasoning of Large Language Model on Knowledge Graph}
\title{Fast Think-on-Graph: Wider, Deeper and Faster \\ Reasoning of Large Language Model on Knowledge Graph}
\author {
    Xujian Liang\textsuperscript{\rm 1,\rm 2},
    Zhaoquan Gu\textsuperscript{\rm 2,\rm 3}\thanks{Corresponding Author}
}
\affiliations {
    \textsuperscript{\rm 1}School of Cyberspace Security, Beijing University Of Posts And Telecommunications, China\\
    \textsuperscript{\rm 2}Department of New Networks, Peng Cheng Laboratory, Shenzhen, China\\
    \textsuperscript{\rm 3}School of Computer Science and Technology, Harbin Institute of Technology (Shenzhen), Shenzhen, China \\
    xjliang@bupt.edu.cn, guzhaoquan@hit.edu.cn
}

\usepackage{bibentry}

\begin{document}

\maketitle

\begin{abstract}
Graph Retrieval Augmented Generation (GRAG) is a novel paradigm that takes the naive RAG system a step further by integrating graph information, such as knowledge graph (KGs), into large-scale language models (LLMs) to mitigate hallucination. However, existing GRAG still encounter limitations: 1) simple paradigms usually fail with the complex problems due to the narrow and shallow correlations capture from KGs 2) methods of strong coupling with KGs tend to be high computation cost and time consuming if the graph is dense. In this paper, we propose the Fast Think-on-Graph (FastToG), an innovative paradigm for enabling LLMs to think ``community by community" within KGs. To do this, FastToG employs community detection for deeper correlation capture and two stages community pruning - coarse and fine pruning for faster retrieval. Furthermore, we also develop two Community-to-Text methods to convert the graph structure of communities into textual form for better understanding by LLMs. Experimental results demonstrate the effectiveness of FastToG, showcasing higher accuracy, faster reasoning, and better explainability compared to the previous works.
\end{abstract}

\begin{links}
\link{Code}{https://github.com/dosonleung/FastToG}
\end{links}

\section{Introduction}

\begin{figure*}[ht!]
  \includegraphics[width=\textwidth]{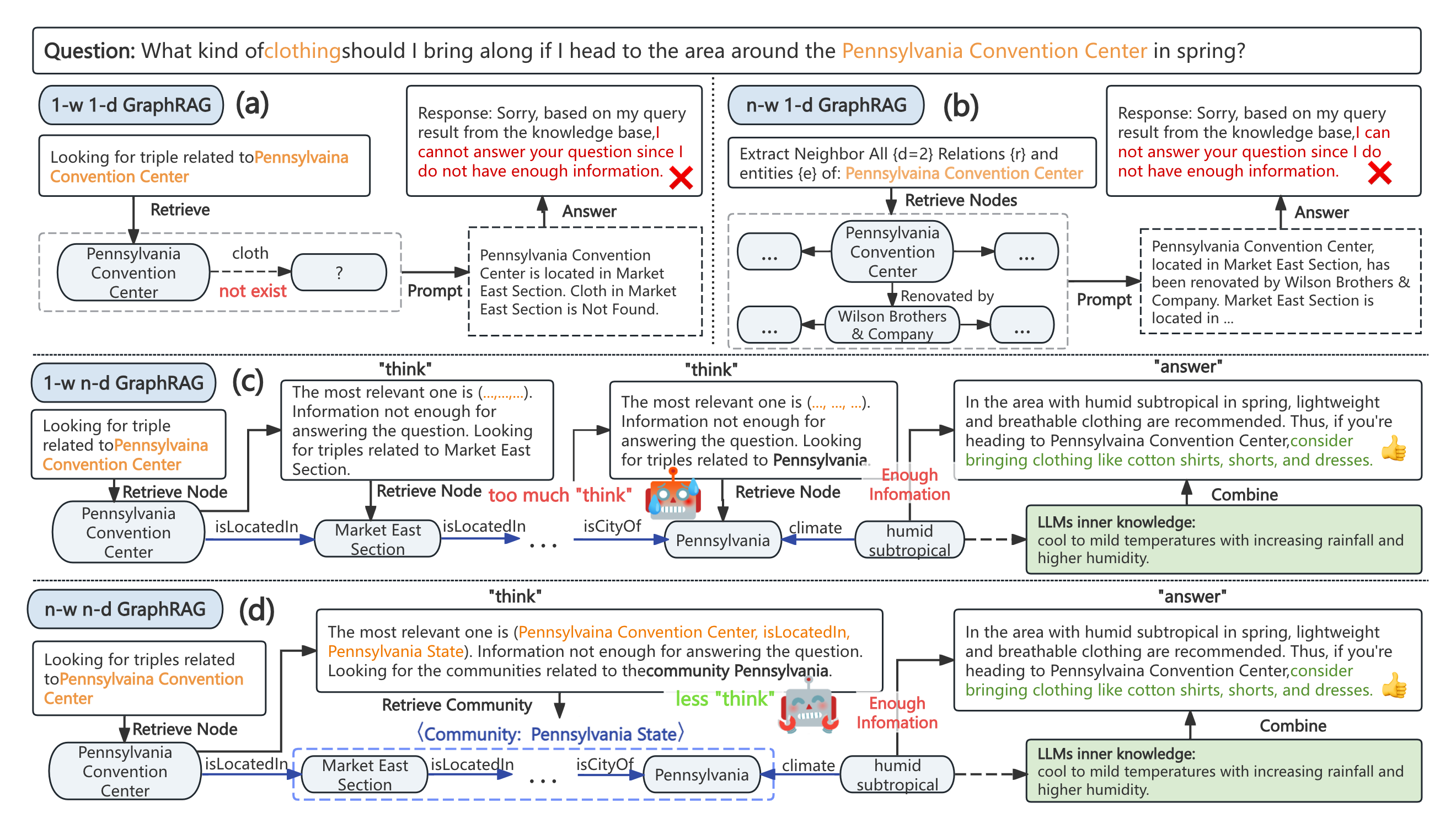}
  \caption{Comparison of 1-w 1-d Graph RAG (a), n-w Graph RAG (b), n-d Graph RAG (c) and n-w n-d GraphRAG (d)}
  \label{fig:main1}
\end{figure*}

Retrieval-Augmented Generation (RAG) \cite{lewis2020retrieval} is a cutting-edge technique for large-scale language models (LLMs) to generate accurate, reasonable, and explainable responses. The early RAG, known as Naive RAG \cite{gao2023retrieval}, mostly works by indexing the documents into manageable chunks, retrieving relevant context based on vector similarity \cite{karpukhin2020dense} to a user query, integrating the context into prompts, and generating the response via LLMs. As the paradigm is simple and highly efficient for the basic tasks, Naive RAG is widely adopted in various applications e.g. Q\&A \cite{chan2024rq}, Recommendation System \cite{deldjoo2024review}, Dialogue System \cite{ni2023recent}, etc. However, there remain such as low precision as the ambiguity or bias \cite{thurnbauer2023towards} exist in the embedding model, low recall when dealing with complex queries, and lack of explainability as the queries and documents are embedded in high-dimensional vector spaces.

The Graph-based RAG (GRAG) is widely considered an advanced RAG by incorporating KGs as an external source for LLMs. Employing various graph algorithms within a graph database, GRAG empowers LLMs for complex operations \cite{10577554} such as BFS and DFS querying, path exploration, and even community detection, etc, providing LLMs a wider and deeper understanding of the relations within the data, making it possible to execute sophisticated reasoning and the generate more precise responses. In this paper, GRAG is categorized into \textbf{n-w GRAG} and \textbf{n-d GRAG} (Fig. \ref{fig:main1}) based on the breadth and depth of retrieval. Particularly, there are still quite a few 1-w 1-d Graph RAG researches \cite{andrus2022enhanced, baek2023knowledge, li2023chain}. Similar to the ``Retrieve one Read One" paradigm of Naive RAG, most of the 1-w 1-d research focuses on single entity or one-hop relationship queries, thereby inheriting the shortcomings of Naive RAG. To overcome these, researches \cite{jiang2023structgpt, modarressi2023ret, edge2024local, wang2024knowledge}, categorized as n-w GRAG, aim to expand the context window of retrieval. Noteworthy works within the n-w GRAG category focus on treating the densely interconnected groups of nodes as the basic units for retrieval and reasoning. These groups are partitioned from the graph by community detection, allowing them to furnish LLMs with more contextual information compared to single node or random clustering methods. For example, given the query concerning the ``climate of the area where Pennsylvania Convention Center is located", triples such as (Pennsylvania Convention Center, located in, Market East Section),...,(Pennsylvania, climate, humid subtropical) would be sufficient for answering as the entity \textit{Pennsylvania Convention Center} of the query is linked to the relevant entity \textit{humid subtropical} over a short distance (n-hop). Despite its simplicity, the retrieval may falter if the distance is too long between the entities (Fig \ref{fig:main1} a,b). n-d GRAG \cite{sun2023think, wang2022knowledge} are ideal for this predicament by broadening the depth of the context window. Analogous to ``Let's think step by step" \cite{kojima2022large}, n-d GRAG takes entities and relations as intermediate steps in the chain of thought (CoT) to guide the generation of LLMs. As they are tightly coupled with KGs, these paradigms exhibit higher recall in retrieval. However, since LLMs are required to ``think" at each step, longer chains result in heightened computational costs (Fig. \ref{fig:main1}, c). On the other hand, if the graph is dense, the multitude of paths between the source and target entities can become exceedingly vast, leading to a decreased recall of entity retrieval.

Building on the consideration of n-w or n-d methods, we propose a \textbf{n-d n-w Graph RAG} paradigm: Fast Think-on-Graph (FastToG). The key idea of FastToG is to guide the LLMs to \textbf{think ``community by community"}. As illustrated in Fig. \ref{fig:main1}, though the path between \textit{Pennsylvania Convention Center} and \textit{humid subtropical} may be lengthy, it notably shortens when nodes are grouped. Accordingly, FastToG regards communities as the steps in the chain of thought, enabling LLMs to ``think" wider, deeper, and faster. Concretely, FastToG leverages community detection on the graph to build the reasoning chains. Considering the time complexity of community detection, we introduce Local Community Search (LCS), aiming to literately detect the communities in a local scope. Given potential graph density concerns, LCS incorporates two stages of community pruning: modularity-based coarse pruning and LLMs-based fine pruning, aiming to enhance the efficiency and accuracy of retrieval. Furthermore, as the language models are trained on textual data, graph structures are incompatible with the format of input. In light of this, We explore two methods to convert community into text: Triple2Text and Graph2Text, which aim to provide better inputs for LLMs.

We conducted experiments on real-world datasets, and FastToG exhibited the following advantages:
\begin{enumerate}
    \item Higher Accuracy: FastToG demonstrate its significant enhancement on the accuracy of LLMs generated content compared with the previous methods.
    \item Faster Reasoning: Our experiments also show that community-based reasoning can notably shorten the reasoning chains, reducing the number of calls to the LLMs.
    \item Better Explainability: The case study indicates that FastToG not only simplifies the retrieval for LLMs but also enhances the explainability for users. 
\end{enumerate}

\begin{figure*}[ht!]
  \centering
  \includegraphics[width=0.96\textwidth]{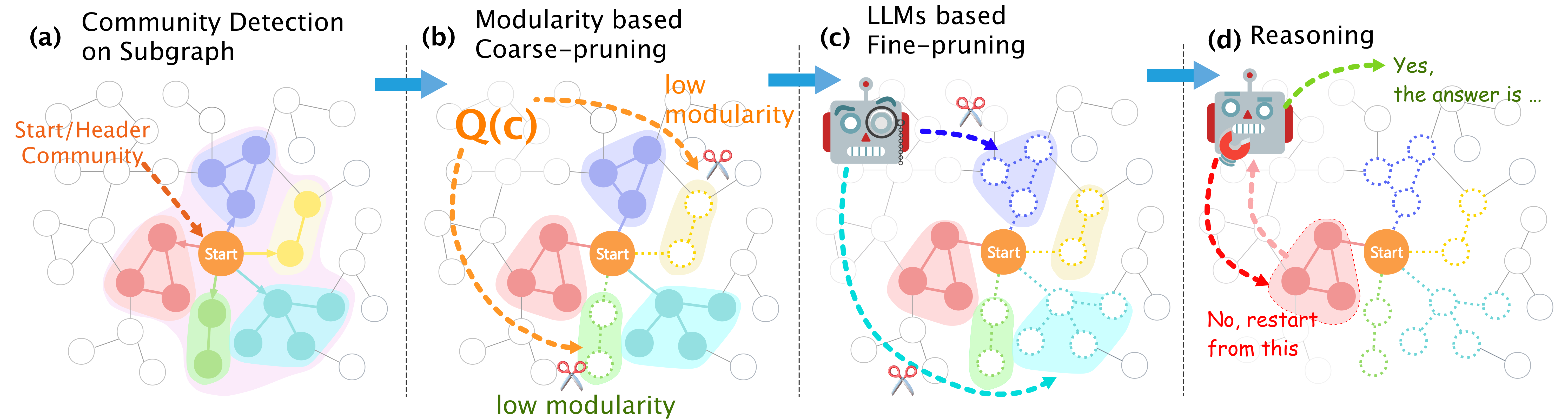}
  \caption{A general schema of the FastToG paradigm.}
  \label{fig:main2}
\end{figure*}

\section{Related Work}
\subsection{Algorithms of Community Detection}
Community detection algorithms are used to grouping or partitioning nodes, as well as their tendency to strengthen or separate apart. These algorithms can be categorized into agglomerative and divisive types.

\textbf{Agglomerative:} These algorithms iteratively group smaller communities until a stop condition is met. One prominent method is hierarchical clustering, which progressively groups nodes or small communities into larger ones by similarity. This down-top grouping procedure is intuitive but computationally expensive, making it impractical for large-scale networks. Since the determination to stop the clustering is important, Louvain \cite{blondel2008fast} and Leiden \cite{traag2019louvain} algorithms introduce the modularity benefit function as a measure of community quality. Based on modularity, the optimization procedure can be explicitly stopped as the global modularity can no longer be improved given perturbations.

\textbf{Divisive:} In contrast, Divisive methods follow the up-bottom process of iteratively partitioning the original graph into small substructures. For example, the Girvan-Newman (GN) algorithm \cite{girvan2002community} removes edges with high betweenness centrality to foster more connected components as communities. Spectral clustering leverages the spectrum(eigenvalues) to embed nodes to k-dimensional vector space and group them based on the clustering method e.g. k-means.

\subsection{Graph-based Retrieval Augmented Generation}
Graph RAG empowers LLMs to capture and utilize the structure and semantics of KGs, enhancing the accuracy of response and explainability of reasoning process. According to the stage at which KGs participate in, GRAG are categorized into Pretrain/Fine-tuning based and Prompt-guided based methods.

\textbf{Pretrain/Fine-tuning based:} These methods enable language models to learn KGs by Input augmentation or model enhancement. \cite{xie2022unifiedskg} linearize graph structures to fine-tune language models, whereas K-BERT expands input structures with KGs information. \cite{zhang2019ernie} embed the entities by incorporating knowledge encodes. \cite{wang2020k} infuses knowledge into pretrained models via transformer adapters, and \cite{yamada2020luke} enhances transformers with an entity-aware self-attention mechanism. While these methods effectively integrate KGs into language models, they require additional supervised data, leading to increasing costs as the model scaling increases.

\textbf{Prompt-guided based} As the training/fine-tuning process is often expensive and time-consuming, recent RAG systems focus on prompt-guided methods. Based on the width and depth of retrieval, these works can be categorized into 1-w 1-d, n-w, and n-d methods. Examples of 1-w 1-d methods include Cok \cite{li2023chain}, KAPING \cite{baek2023knowledge} etc. Cok enhances the flexibility of queries with SPARQL. KAPING vectorizes triples for efficient retrieval. Representative works of n-w includes KGP \cite{wang2024knowledge}, StructGPT \cite{jiang2023structgpt}, RET-LLMs \cite{modarressi2023ret} and GraphRAG \cite{edge2024local} etc. While the first three all randomly or fully capture the adjacent nodes for retrieval purposes, GraphRAG stands out by utilizing community detection to optimize retrieval for text summarization. KP-PLM \cite{wang2022knowledge} and ToG \cite{sun2023think} are the typical works of the n-d paradigm. KP-PLM firstly designs the $d>1$ entity-relation chains for information retrieval. ToG, particularly relevant to our research, introduces enhancements such as pruning and beam search within chains via LLMs.

\section{The Method}

\subsection{Overview}

We begin by introducing the basic framework of Fast think on graph (FastToG). Inspired by the work \cite{kojima2022large} ``Let's think step by step", the basic idea of FastToG is enabling large models to ``think" community by community on the KGs. Consequently, the core component of the FastToG (see our repository for the pseudocode) are the $W$ reasoning chains $P = $[$p_1$, $p_2$, ..., $p_W$], where each chain $p_i \subseteq P$ is a linked list of communities i.e., $p_i = $ [$c_0^i$, $c_1^i$, ..., $c_{n-1}^i$].

The FastToG framework comprises two main phases: the initial phase and the reasoning phase. The objective of the initial phase is to determine the start communities $c_0$ and the header community (Fig. \ref{fig:main2}a) $c_0^i$ for each reasoning chain $p_i$. FastToG prompts LLMs to extract the subject entities of to query $x$ as a single-node community $c_0$. After that, FastToG employs Local Community Search (LCS), which involves two key components: Community Detection on Subgraph (Fig. \ref{fig:main2}b) and Community Pruning (Fig. \ref{fig:main2}c), to identify neighbor communities with highest potential for solving the query $x$, serving as the head community $c_0^i$ for each reasoning chain $p_i$.

Once the head communities are determined, the algorithm enters into the reasoning phase. For each $p_i$, FastToG continues to leverage LCS for local community detection and pruning, with the selected community being added to $p_i$ as the newest element, termed \textbf{pruning}. After the update of all reasoning chains, Community2Text methods like Graph2Text (G2T) or Triple2Text (T2T) are utilized to convert all communities within chains into textual form as the input content of LLMs, which is called \textbf{reasoning} (Fig. \ref{fig:main2}d). If the gathered reasoning chains are deemed adequate by the LLMs for generating the answer, the algorithm will be terminated and return the answer. To mitigate time consumption, if no answer is returned within $D_{max}$ iterations of reasoning, the algorithm will be degraded into the methods of answering the query by inner knowledge of LLMs e.g. IO/CoT/CoT-SC. We will illustrate all the details in the following sections.

\subsection{Local Community Search}

In this subsection, we focus on the LCS, which consist of there main part: community detection on the subgraph, pruning methods of coarse-pruning and fine-pruning.

\subsubsection{Community Detection on Subgraph}

Due to the vast number of nodes and relations in the KGs, as well as the dynamic of queries, it is impossible to partition the KGs into communities entirely and frequently. To solve this, we propose community detection based on the local subgraphs for each pruning. Given KGs $\Omega$, start community $c_0$  or header community $c_0^i$, FastToG firstly retrieves the subgraph $g$ ($g \subset \Omega$) within $n$ hops from $c_0$ or $c_0^i$. Considering the rapid growth of the number of neighbors in the dense graph and the degradation of semantics as the nodes move further apart, the algorithm randomly samples neighbor nodes at different hops with exponential decay $\rho$. The probability of selecting each node $x$ at $n$-hop from $c_0$ or $c_1^i$ is given by:
$$Pr(x=1)=\rho^{n-1}$$

Once the local subgraph $g$ is retrieved, the next step is to partition it into communities. Community detection algorithms are utilized to reveal the closely connected groups of nodes (referred to as communities) of the graph. This process enables the graph to be partitioned into subgroups with tight internal connectivity and sparse external connections, aiding LLMs in uncovering implicit patterns within the graph structure. In this study, we consider Louvain \cite{blondel2008fast}, Girvan–Newman algorithm \cite{girvan2002community}, Agglomerative Hierarchical Clustering, and Spectral Clustering as representative algorithms. To prevent the retrieval of repeated communities, the algorithm ensures that new communities discovered do not exist in the historical community set during each community detection iteration, and adds them to the historical community set after each pruning step.

The oversize communities may contain more redundant nodes, leading to the increased noise, prolonged community detection time, and reduced explainability. Thus, we introduce a constraint on the maximum size of community within the community detection. To do this, all detection algorithms will be backtracked when the stop condition is met. At each iteration from the end to the beginning of backtracking, the algorithm verifies whether the size of each community meets the condition $\text{size}(c_i) <= M$ (hyperparameter). Finally, the algorithm returns the partition status that first satisfies the size constraint.

\subsubsection{Modularity-based Coarse-Pruning}

When dealing with too many candidate communities, existing methods relying on Large Language Models (LLMs) for community selection encounter difficulties, resulting in either high time consumption or selection bias \cite{zheng2023large}. To tackle this, current methods frequently incorporate random sampling to diminish the number of candidate communities or nodes. However, these methods inevitably ignore the network structure of the graph, resulting in the loss of structural information from the candidate communities.

Based on the above, we propose modularity-based coarse-pruning. The modularity \cite{blondel2008fast} of a partitioned subgraph $g$ is:

\begin{align}
    Q=\frac{1}{2m}\sum_{ij}[A_{ij}-\frac{k_ik_j}{2m}] \delta (c_i,c_j)
\label{fun:modularity1}
\end{align}

Where $m$ is the number of edges, $A_{ij}$ is an element in the adjacency matrix $A$, $k_i$ and $k_j$ are the degrees of nodes $i$ and $j$, respectively. The function $\delta(c_i, c_j) = 1$ indicates that nodes $i$ and $j$ belong to the same community, otherwise, it returns 0. In the weighted graph, $m$ is the sum of graph weights, $\frac{1}{2m}\sum_{ij}A_{ij}\delta (c_i,c_j)$ represent the average weight of the given community, while $\frac{k_ik_j}{2m}\delta (c_i,c_j)$ denotes the average weight of the given community under random conditions. The larger the difference between them, the higher the quality of community partitioning. For simplicity, we do not consider the weights or directions of edges in our work. The expression\ref{fun:modularity1} of modularity can be rewritten as:

\begin{align}
    Q &= \frac{1}{2m}[\sum_{ij}A_{ij}-\frac{\sum_ik_i\sum_jk_j}{2m}] \delta (c_i,c_j) \\
      &= \frac{1}{2m}\sum_c[\sum \text{in}-\frac{(\sum \text{tot})^2}{2m}]
\end{align}

where $\sum \text{in}$ is the number of edges in community $c$ and $\sum \text{tot}$ is the number of edges connected to $c$. Thus, the modularity of community $c$ can be:

\begin{align}
    Q(c)=\sum \text{in} - \frac{(\sum \text{tot})^2}{2m}
\end{align}

Upon calculating the modularity of each candidate community $c \subseteq C$, the communities $C^{'}$ with low modularity will be pruned, while the remaining will serve as the refined set of candidate communities for the next stage.

\begin{align}
    C^{'} := \text{argtopk}_{c \subseteq C} Q(c)
\end{align}

\subsubsection{LLMs-based Fine-Pruning}
After coarse pruning, a smaller set of candidate communities $C^{'}$ that are more compact in structure is identified. Subsequently, FastToG prompts the LLMs to make the final selection $C^{''}$:

\begin{align}
C^{''} = \text{fine\_pruning}(x, C^{'}, \Pi, k)
\end{align}

where $x$ is the query, $\Pi$ is the instance of LLMs, $k=1$ or $W$ is for the single or multiple choice.

To simplify the pruning process, FastToG no longer considers scoring-based pruning \cite{sun2023think}. Instead, it prompts the LLMs to directly choose either the best community or top $W$ communities. In the initial phase, LLMs are guided through multiple-choice prompts to retrieve $W$ communities, which will act as the header communities $c_0^1$,...,$c_0^W$ for the reasoning chains $p_1$,...,$p_W$, respectively. During the reasoning phase, single-choice prompts are employed for each reasoning chain to recall the best community $c_j^i$, which is then appended to its chain $p_i$. Not that the length of each chain $p$ may not be the same if LLMs insist that none of the candidate communities are relevant to the query. In such cases, the exploration of such chains will be discontinued.

\subsection{Reasoning}

Once of all the reasoning chains $p_i \subset P$ are updated, LLMs will be prompted to integrate and generate the answers from all the chains. The returned results could be a clear answer if the chains are adequate for answering, or ``Unknown" if not. In cases where ``Unknown" is returned, the algorithm proceeds to the next iteration until either reaching the maximum depth $D=D_{max}$ or obtaining a definitive answer. If the maximum iterations are exhausted without a conclusive response, FastToG will be degraded to generate the answer by the inner knowledge of LLMs itself. Consistent with ToG \cite{sun2023think}, the entire process of FastToG paradigm involves 1 round of pruning and reasoning in the initial phase, and $2WD_{max}$ pruning and $D_{max}$ reasoning in the second. Consequently, the worst condition needs $2WD_{max} + D_{max} + 2$ calls to the LLMs.  

\subsection{Community-to-Text}
Knowledge graphs are organized and stored in the form of RDF triples. Thus, a community also consisted of triples e.g. $c$ = [(Philadelphia, isCityOf, Pennsylvania), (Pennsylvania, climate, Humid Subtropical)]. To input this structure into LLMs, it needs to be converted into text format. To do this, we propose two methods: Triple-to-Text (T2T) and Graph-to-Text (G2T). For T2T, triples are directly converted into text by rule-based scripts e.g. T2T($c$) = ``Philadelphia isCityOf Pennsylvania, Pennsylvania climate Humid Subtropical". For G2T, triple will be converted into human language like G2T($c$)  = ``Philadelphia, located in the state of Pennsylvania, features a Humid Subtropical climate."

T2T may result in redundancy. For instance, a T2T result of [(Allentown, isCityOf, Pennsylvania), (New Castle, isCityof, Pennsylvania), (Philadelphia, isCityOf, Pennsylvania)] is ``Allentown isCityOf Pennsylvania, New Castle isCityof Pennsylvania, Philadelphia isCityOf Pennsylvania", which can be summarized as: ``Allentown, New Castle, and Philadelphia are the cities of Pennsylvania". Therefore, G2T also undertakes the role of the text summary. To do this, we fine-tune the smaller language models (like T5-base, 220M) on the outputs from T2T for the conversion.

Since G2T leverages a base model with less parameters compared to existing LLMs (e.g., llama-3-8b has 36 times more parameters than T5-base), the impact of time efficiency is tiny. Apart from intra-community relationships, there also exist inter-community paths such as $c_1 - E_{1,2} - c_2$. Therefore, it is necessary to perform text transform on the $E_{1,2}$. This study excludes candidate communities with distance larger than 1 hop from the current community, meaning the paths between the current community and candidate communities do not contain any intermediate nodes.

\section{Experiments}

\subsection{Experimental Setup}
\subsubsection{Dataset and Evaluation Metric} 
We evaluated FastToG on 6 real-world datasets, which include datasets of multi-hop KBQA: CWQ \cite{talmor-berant-2018-web}, WebQSP \cite{yih-etal-2016-value}, and QALD \cite{perevalov2022qald}, slot filling: Zero-Shot RE (abbr. ZSRE) \cite{petroni-etal-2021-kilt} and TREx \cite{ELSAHAR18.632}, and common-sense reasoning Creak \cite{onoe2021creak}. To ensure comparability with prior works \cite{li2023chain, sun2023think, edge2024local}, we employed exact match (hit@1) as the evaluation metric. Considering the computational cost and time consumption, we randomly sampled 1k examples from the datasets containing more than a thousand data. Furthermore, the initial entities provided by \cite{sun2023think} were used as the starting community for the evaluations.

\subsubsection{Language Model} 
To reduce the computational cost, we employ two LLMs: gpt-4o-mini\footnote{https://openai.com/index/gpt-4o-mini-advancing-cost-efficient-intelligence} and Llama-3-70b-instruct \cite{dubey2024llama} without quantization. To keep the pruning steady, the temperature is configured to 0.4 during pruning and 0.1 during reasoning. Considering the text descriptions of community structure are longer, the maximum length of output is increased to 1024.

\subsubsection{Graph2Text Model}
The Graph2Text module is developed by fine-tuning T5-base \cite{raffel2020exploring} using dataset WebNLG \cite{auer2007dbpedia} and synthetic data generated by GPT-4. To build the synthetic data, we prompt GPT-4 to generate text descriptions of given communities, which are the byproduct of the experiments.

\subsubsection{Knowledge Graph}
We utilize Wikidata \cite{vrandevcic2014wikidata}, which is a free and structured knowledge base, as source of KGs. The version of Wikidata is 20240101 and only the triples in English are considered. Following extraction, 84.6 million entities and 303.6 million relationships are stored in Neo4j\footnote{https://neo4j.com/}.

\subsection{Performance on Accuracy}
Accuracy is one of the most important criteria for RAG systems. In this experiment, we evaluate FastToG and compared methods on the datasets and settings mentioned above.   

\begin{table}[!h]
\footnotesize
\centering
\begin{tabular}{lllllll}
\hline
\multicolumn{1}{c}{\scriptsize \textbf{Method}} & \multicolumn{1}{c}{\scriptsize \textbf{CWQ}} & \multicolumn{1}{c}{\scriptsize \textbf{WebQSP}} & \multicolumn{1}{c}{\scriptsize \textbf{QALD}} & \multicolumn{1}{c}{\scriptsize \textbf{ZSRE}} & \multicolumn{1}{c}{\scriptsize \textbf{TREx}} & \multicolumn{1}{c}{\scriptsize \textbf{Creak}} \\ \hline
\multicolumn{7}{c}{\scriptsize  Inner-knowledge based Methods}                                                                                                                                                            \\ \hline
IO           & 31.2  & 49.6  & 38.6  & 26.4  & 46.4  & 90.2                      \\
CoT          & 35.1  & 60.8  & 51.8  & 35.6  & 52.0  & 94.6                      \\
CoT-SC       & 36.3  & 61.2  & 52.4  & 35.8  & 52.0  & 95.0                      \\ \hline
\multicolumn{7}{c}{\scriptsize  KGs-retrieval based Methods}                                                                                                                                                              \\ \hline
1-d 1-w      & 35.5  & 59.2  & 50.7  & 39.4  & 56.1  & 92.0                         \\
1-d n-w      & 42.3  & 64.4  & 54.8  & 46.1  & 58.8  & 92.8                         \\
n-d 1-w      & 42.9  & 63.6  & 54.9  & 54.0  & 64.2  & 95.4                      \\
Ours(t2t)    & 43.8  & 65.2  & \textbf{56.1}  & \textbf{54.4}  & 67.3  & 95.6                      \\
Ours(g2t)    & \textbf{45.0} & \textbf{65.8}  & 55.9  & 54.2  & \textbf{68.6}  & \textbf{96.0}                      \\ \hline
\end{tabular}
\caption{Accuracy (\%) for different datasets by gpt-4o-mini.}
\label{table:main_gpt4omini}
\end{table}

\begin{table}[!h]
\footnotesize
\centering
\begin{tabular}{lllllll}
\hline
\multicolumn{1}{c}{\scriptsize \textbf{Method}} & \multicolumn{1}{c}{\scriptsize \textbf{CWQ}} & \multicolumn{1}{c}{\scriptsize \textbf{WebQSP}} & \multicolumn{1}{c}{\scriptsize \textbf{QALD}} & \multicolumn{1}{c}{\scriptsize \textbf{ZSRE}} & \multicolumn{1}{c}{\scriptsize \textbf{TREx}} & \multicolumn{1}{c}{\scriptsize \textbf{Creak}} \\ \hline
\multicolumn{7}{c}{\scriptsize Inner-knowledge based Methods}                                                                                                                                                            \\ \hline
IO           & 32.9  & 55.8  & 44.3  & 30.7  & 52.8  & 91.0                      \\
CoT          & 34.2  & 58.8  & 45.0  & 32.6  & 55.6  & 91.2                      \\
CoT-SC       & 34.8  & 60.6  & 46.8  & 34.9  & 57.0  & 91.8                      \\ \hline
\multicolumn{7}{c}{\scriptsize KGs-retrieval based Methods}                                                                                                                                                              \\ \hline
1-d 1-w      & 35.0  & 57.6  & 44.4  & 34.4  & 56.0  & 91.3                      \\
1-d n-w      & 39.8  & 64.0  & 46.6  & 57.1  & 60.4  & 91.9                      \\
n-d 1-w      & 40.3  & 62.4  & 51.6  & 64.8  & 61.2  & 93.3                      \\
Ours(t2t)    & 42.1  & 65.9  & \textbf{54.9}  & 67.7  & 63.5  & 92.2                         \\
Ours(g2t)    & \textbf{46.2}  & \textbf{66.4}  & 54.3  & \textbf{67.9} & \textbf{64.7}  & \textbf{94.5}                         \\ \hline
\end{tabular}
\caption{Accuracy (\%) for different datasets by llama-3-70b.}
\label{table:main_llama3}
\end{table}

\subsubsection{Compared Methods} We consider two categories of comparative methods: Inner-knowledge-based or KGs-retrieval-based methods. The former methods include: 1) IO prompt \cite{brown2020language}: prompting the model to directly generate the result. 2) CoT \cite{wei2022chain}: guiding the model to ``think" step by step before answering. 3) CoT-SC \cite{wang2022self}: ensembling all the CoTs to obtain more consistent predictions. The KGs-retrieval-based methods include: 1) 1-d 1-w methods, which represent the methods like CoK \cite{li2023chain}, KAPING \cite{baek2023knowledge}, etc. 2) 1-d n-w methods for KGP \cite{wang2024knowledge} 3. n-d 1-w method for ToG \cite{sun2023think}. To keep consistency with previous works, we keep a $W=3$ for the number of chains and $D_{max}=5$ for the maximum iterations. For n-w methods like 1-d n-w, Ours(t2t), and Ours(g2t), the maximum size of community is all set at 4 and the algorithm for community detection are Louvain Method.

\subsubsection{Compared Result}
Tab. \ref{table:main_gpt4omini} and Tab. \ref{table:main_llama3} respectively display the accuracy achieved by gpt-4o-mini and llama3-70b-instruct across the datasets. Overall, FastToG, which includes t2t and g2t mode, outperforms all previous methods. In particular, Ours(g2t) surpasses n-d 1-w (ToG) by 4.4\% in Tab. \ref{table:main_gpt4omini} and 5.9\% in Tab. \ref{table:main_llama3}.

For two community-to-text conversions, g2t methods show higher accuracy on most datasets, aligning well with the idea for our proposed Graph2Text in the previous section. However, we note that the improvement of g2t is tiny (mostly $< 1\%$), and even slightly underperforms to t2t on QALD dataset. To find out the explanation for these counterintuitive results, we carried out extensive checks (see our repository for details) and found that the hallucination from the base model of Graph2Text is the main reason for g2t falling short of t2t.


\begin{figure}[t!]
\centering
\includegraphics[width=0.95\columnwidth]{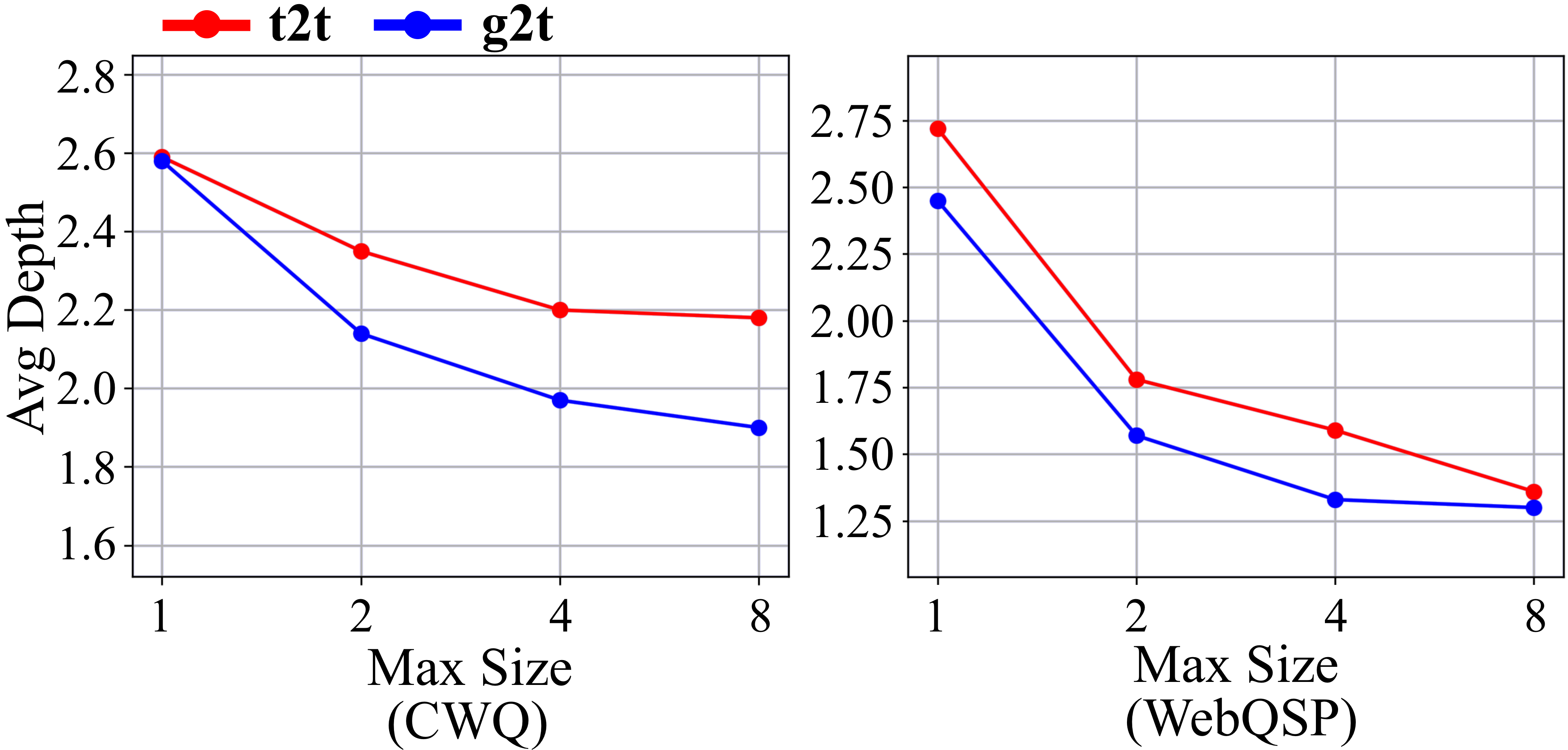}
\caption{Average Depth versus Max size of community}
\label{main_depth}
\end{figure}

\begin{figure}[t!]
\centering
\includegraphics[width=0.95\columnwidth]{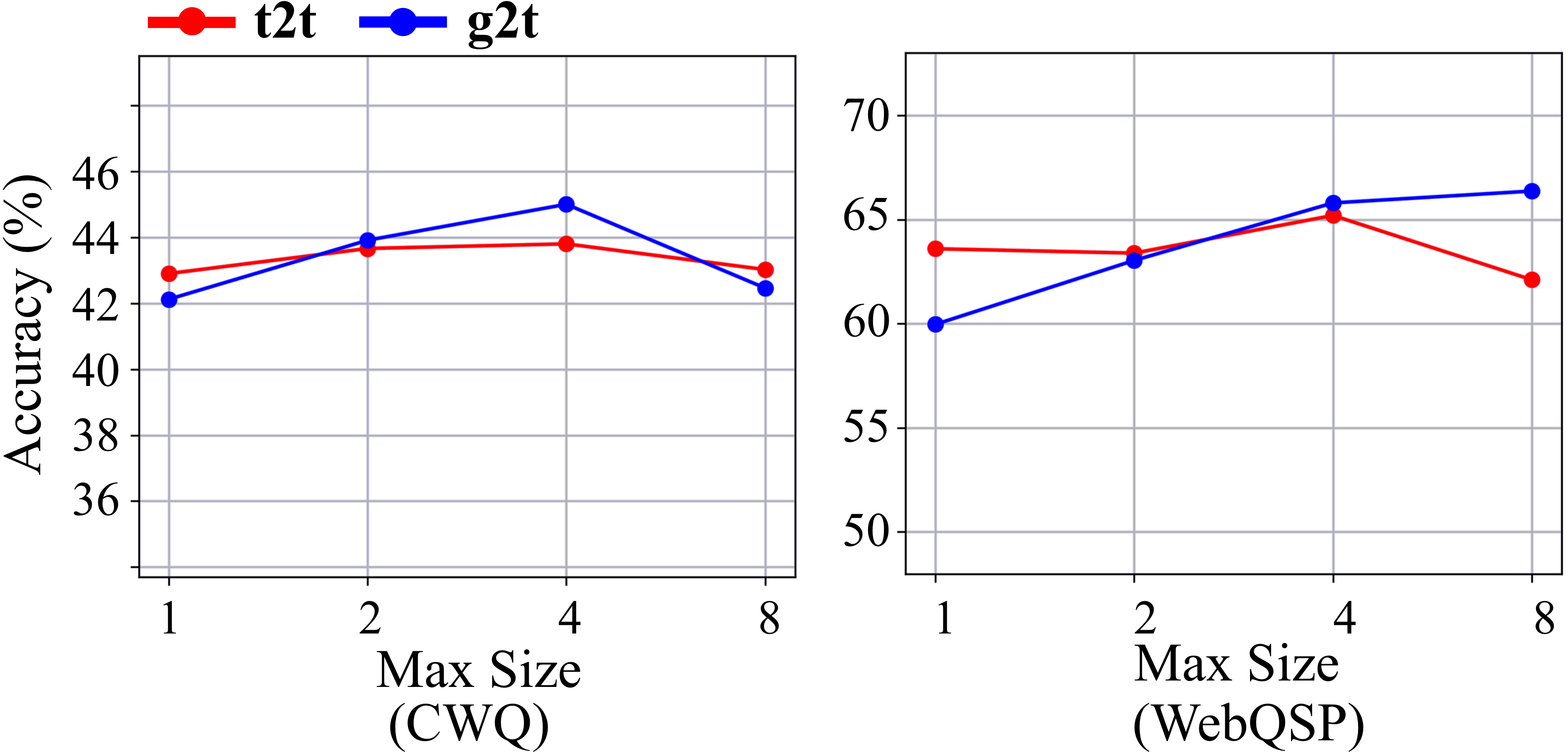}
\caption{Accuracy versus Max size of community}
\label{ablation_acc}
\end{figure}

\begin{figure}[t!]
\centering
\includegraphics[width=1.0\columnwidth]{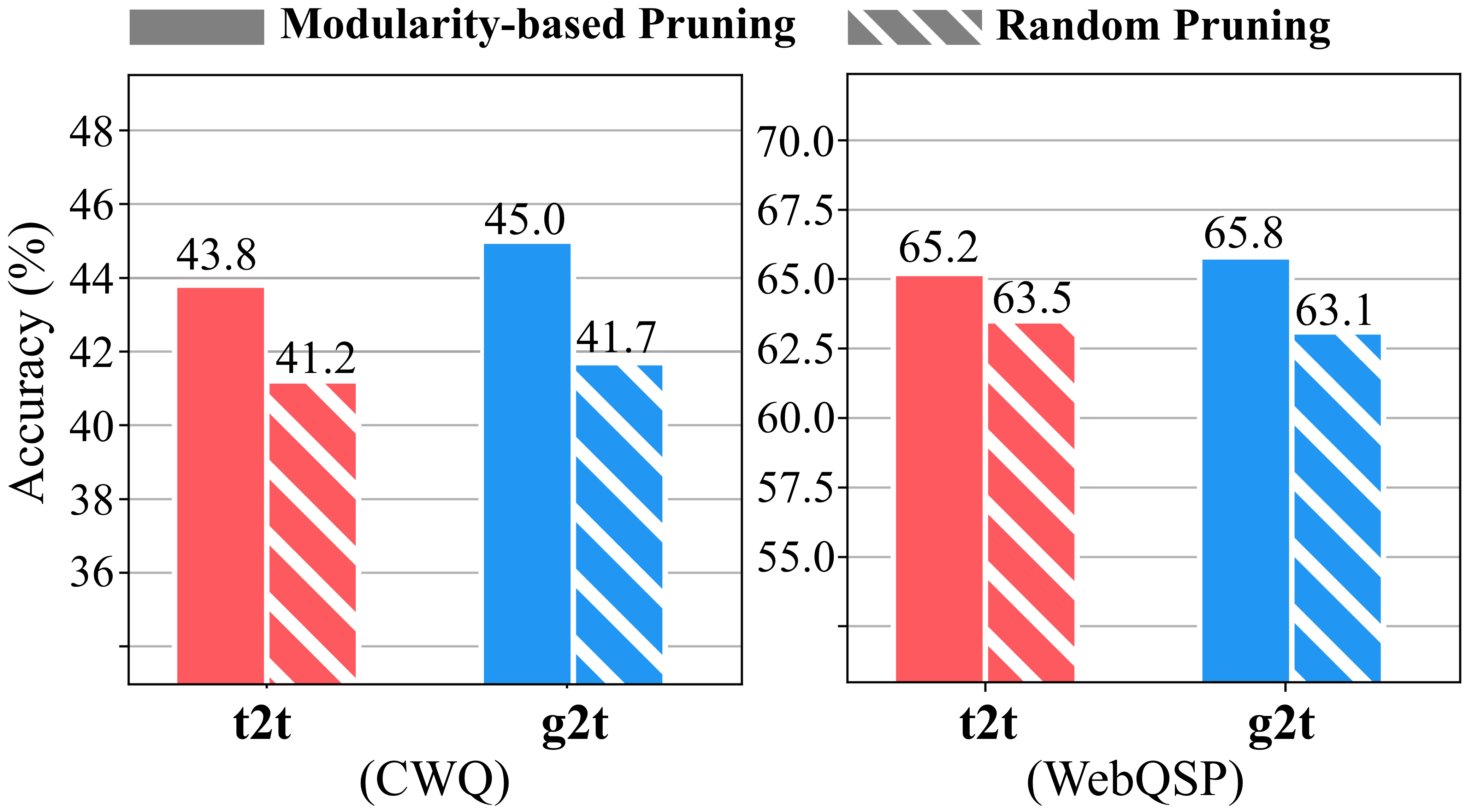}
\caption{Accuracy (\%) of two pruning methods}
\label{ablation_pruning}
\end{figure}

\subsection{Performance on Efficiency}

Efficiency is another key criterion of our work. To verify it, we consider the number of calls to the LLMs as the evaluation criterion. As outlined in Section Reasoning, for each question, the number of calls to the LLMs is estimated as $2WD + D + 2$. Since all the settings of $n-d$ methods have the same W, we only need to compare $D$ to for the evaluation as efficiency $\epsilon \propto D$. Methods IO, CoT, CoT-SC, and 1-d are not considered because the value of D is fixed at 1.

Fig. \ref{main_depth} illustrates the relationship between the max size of the community and average depth on the CWQ and WebQSP datasets, where $Max Size = 1$ refers to the previous n-d 1-w works like ToG. It is evident that FastToG ($Max Size > 1$) outperforms the n-d 1-w methods. Even with the communities with $Max Size = 2$, there can be a significant reduction in the $Avg Depth$ of the reasoning chains. For example, the $Avg Depth$ on CWQ is reduced by approximately 0.2 and 0.4 for t2t/g2t modes, respectively. On WebQSP, these reductions increase to about 1.0 for both modes. Moreover, as the $Max Size$ increases, the decrease in depth becomes less substantial. 

\subsection{Ablation Study}
\subsubsection{Trade-off between Accuracy and Efficiency}
While larger communities can reduce the average depth of reasoning chains, such large communities may also bring more noise, potentially negatively impacting the reasoning effectiveness of LLMs. Fig. \ref{ablation_acc} illustrates the relationship between the $Max Size$ and accuracy. Overall, most of the cases achieve higher accuracy when $Max Size$ is set to 4. However, at $Max Size=8$, results show a decrease in accuracy. Therefore, setting a larger size of the community does not necessarily result in higher gain from accuracy. 

\subsubsection{Comparison on Pruning Methods}

To validate the effectiveness of our proposed Modularity-based Coarse Pruning, we compared it with Random Pruning, a method widely used by previous works. Fig. \ref{ablation_pruning} depicts the accuracy comparison between Modularity-based coarse pruning and random pruning on the CWQ and WebQSP datasets for 2 modes, with all cases using $Max Size=4$. Overall, Modularity-based Coarse Pruning outperforms Random Pruning in all cases. Particularly, we observed that cases based on g2t mode are more sensitive in modularity-based pruning, indicating that communities of densely connected structure are preferable for the conversion between graphs and text.

\subsubsection{Comparison on different Community Detection}
Community detection is a crucial step in FastToG. Tab. \ref{table:ablation_community_detection} compares the impact of different detection algorithms — including Louvain, Girvan-Newman (abbr. GN), hierarchical clustering (abbr. Hier), and Spectral Clustering (abbr. Spectral) on accuracy. Additionally, we consider random community detection (abbr. Rand), which randomly partitions nodes into different groups. Comparing the 4 non-random algorithms, the impact of different community detection on FastToG is very small ($<1\%$ on average). On the other hand, when comparing algorithms between random and non-random, the latter outperforms the former ($>3\%$ on average), demonstrating that community detection does help FastToG.

\begin{table}[!t]
\footnotesize
\begin{tabular}{lllllll}
\hline
\multicolumn{1}{l}{}    &     & Rand & Louvain & GN   & Hier & Spectral \\ \hline
\multirow{2}{*}{\scriptsize \textbf{CWQ}}    & t2t & 40.4 & 43.8    & 44.0 & 44.5 & 44.0      \\
                        & g2t & 42.6 & 45.0    & 45.7 & 46.0 & 45.3      \\ \hline
\multirow{2}{*}{\scriptsize \textbf{WebQSP}} & t2t & 60.6 & 65.2    & 66.8 & 66.0 & 65.1      \\
                        & g2t & 62.5 & 65.8    & 66.9 & 66.1 & 66.7      \\ \hline
\end{tabular}
\caption{Accuracy (\%) of Community Detection Algorithms}
\label{table:ablation_community_detection}
\end{table}

\begin{figure}[t!]
\centering
\includegraphics[width=0.95\columnwidth]{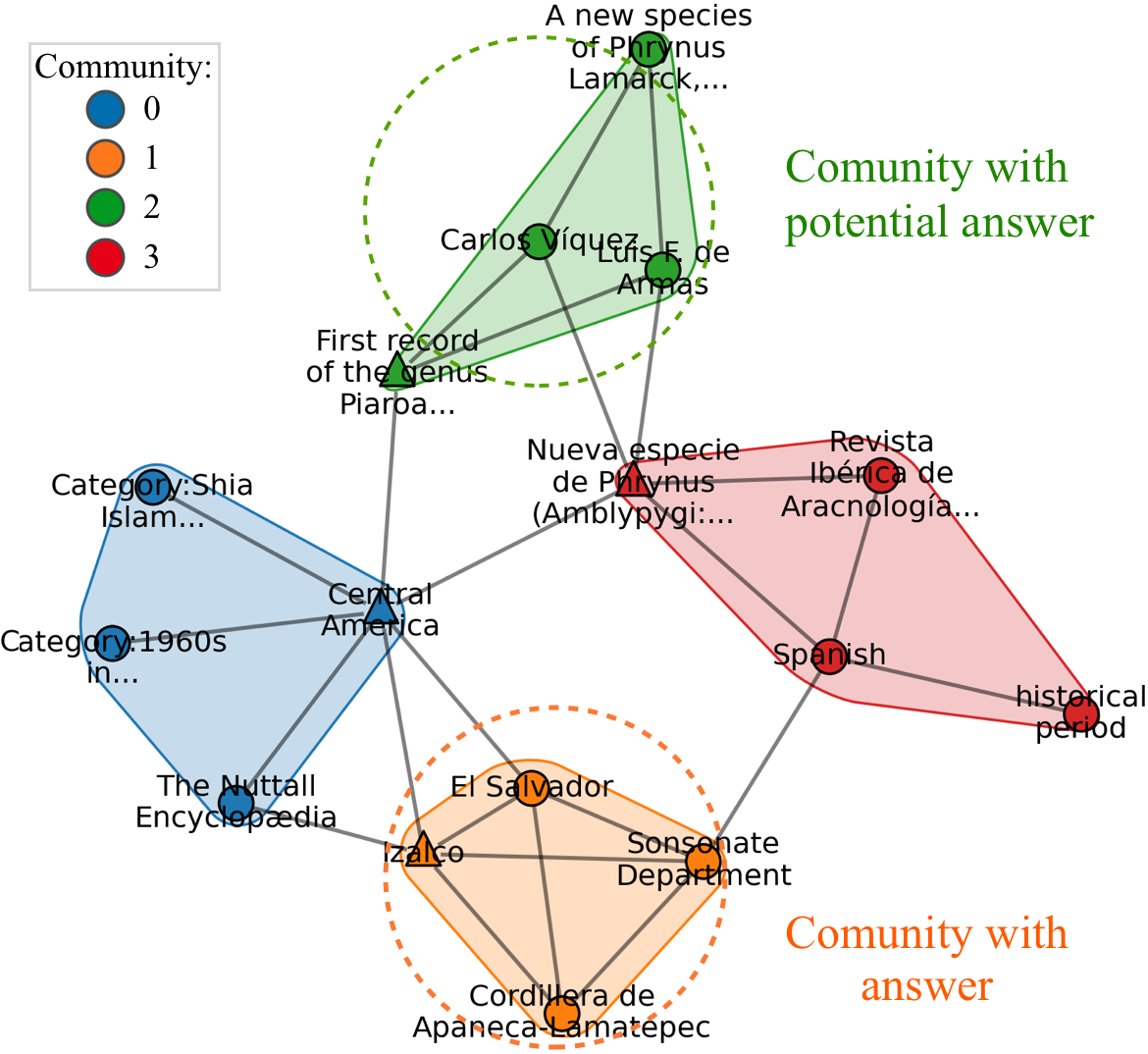}
\caption{Visualization of the retrieval by FastToG}
\label{case_study}
\end{figure}

\begin{table}[]
\scriptsize
\begin{tabular}{ll}
\hline
\multicolumn{1}{c}{\textbf{Think on Community}}                                                                                                                                                                                    & \multicolumn{1}{c}{\textbf{Think on Node}}                                                                                                  \\ \hline
\begin{tabular}[c]{@{}l@{}}A. The Sonsonate Department is located\\  in \textcolor{orange}{El Salvador}, which is part of Central\\  America. \textcolor{red}{Spanish} is the language used in\\  the Sonsonate Department ...\end{tabular}                             & \begin{tabular}[c]{@{}l@{}}A. El Salvador is belong to\\ Central America\end{tabular}                                                      \\ \hline
\begin{tabular}[c]{@{}l@{}}B. The Nuttall Encyclopida describes\\  Mexico City as a city in Central America,\\  which is part of North America. The\\  Centralist Republic of Mexico ...\end{tabular}                              & \begin{tabular}[c]{@{}l@{}}B. Nueva especie de Phrynus is \\ main subject in Central America.\end{tabular}                                  \\ \hline
\begin{tabular}[c]{@{}l@{}}C. The first record of genus Piaroa\\  Villarreal, Giupponi \& Tourinho, 2008, is\\  from Central America. Carlos Vquez, a\\  \textcolor{SeaGreen}{Panamanian} author, has written about ...\end{tabular} & \begin{tabular}[c]{@{}l@{}}... more than 20 options\\ Z. The South and Central Ameri-\\can Club is held in Central \\ America.\end{tabular} \\ \hline
\end{tabular}
\caption{Think on Community versus Think on Node}
\label{table:case_study}
\end{table}

\subsection{Case Study}
For the query ``Of the 7 countries in Central America, which consider Spanish an official language?" from dataset CWQ, we visualize the retrieval process of FastToG. Fig. \ref{case_study} displays a snapshot of LLMs pruning with start community \textit{Central America}. Note that the node \textit{Central America} has 267 one-hop neighbors, making it hard to visualize. Tab. \ref{table:case_study} shows the corresponding Graph2Text output of the communities or nodes. The left column shows part of the pruning with communities as the units to be selected, while the right column shows the nodes. As we can see, community as the basic unity for pruning greatly reduces the number of unit. In addition, Community as the unit can increase the likelihood of solving this problem. For example, comparing two option A, the community associated with the option A in left column has a path connecting to the entity \textit{Spanish}. Thus, we can sure that \textit{EI Salvador} is one of the answers. In contrast, when using Node as the unit, it requires more exploration to reach this entity. Hence, ``think" on the community not only simplifies the pruning process for LLMs but also enhances clarity for users in understanding the reasoning process.

\section{Conclusion}
We introduced a novel GraphRAG paradigm - FastToG. By allowing large models to think ``community by community" on the knowledge graphs, FastToG not only improves the accuracy of answers generated by LLMs but also enhances the efficiency of the RAG system. We also identified areas for further improvement, such as incorporating semantics and syntax with community detection, introducing multi-level hierarchical communities to enhance retrieval efficiency, and developing better community-to-text conversion.

\section{Acknowledgments}
This work was supported in part by the Major Key Project of PCL (No. PCL2024A05), National Natural Science Foundation of China (Grant No. 62372137), Shenzhen Science and Technology Program (No. KJZD20231023094701003), and the opening project of science and technology on communication networks laboratory (No. FFX24641X007).

\bibliography{references}

\newpage
\appendix
\section{Appendix}

\subsection{A. Algorithms for FastToG}
we provide the relevant algorithms for FastToG. Algorithm 1 is the basic algorithmic framework of FastToG and Algorithm 2 is the implementation of Local Community Search.

\begin{algorithm}[h!]
\caption{FastToG}
\label{alg:algorithm1}
\textbf{Require}: LLMs $\pi$, Knowledge Graph $\Omega$\\
\textbf{Input}: query $x$ \\
\textbf{Parameter}: \\
$W$ width of reasoning chains \\
$I_{max}$ maximum iteration for reasoning\\
$R_{max}$ maximum radius of subgraph \\
\textbf{Initialize:} \\
reasoning chains $P \leftarrow $ [] $ \times W$  \\
history set of community $H$ \\
$I \leftarrow 0$ \\
// initial phase
\begin{algorithmic}[1] 
\STATE $c^0 \leftarrow$ extract entities on $x$ \\
\STATE $C^* \leftarrow$ LCS($x$, $c^0$, $\pi$,$\Omega$, $R_{max}$, $W$) \\
\STATE Append each $c^*$ of $C^*$ to $p_i$ of $P$ respectively
\end{algorithmic}
// reasoning phase
\begin{algorithmic}[1] 
\STATE $I = 0$
\WHILE{$I < I_{max}$}
\FOR{each $p_i$ in $P$}
\STATE $c_{I}^i \leftarrow$ get the last community from $p_i$
\STATE $c^* \leftarrow$ LCS($x$, $c_{I}^i$, $\pi$, $\Omega$, $R_{max}$, 1)
\STATE append $c^*$ to $p_i$
\STATE add $c^*$ to $H$
\ENDFOR
\STATE $P_{text} \leftarrow $ Community2Text($P$) 
\IF{reasoning($\pi$, $x$, $P_{text}$)}
\STATE break
\ENDIF
\STATE Increment $I$ by 1
\ENDWHILE
\IF{$I \leq I_{max}$}
\STATE reasoning($\pi$, $x$) //IO,CoT
\ENDIF
\end{algorithmic}
\end{algorithm}

\begin{algorithm}[h!]
\caption{LocalCommunitySearch (abbr. LCS)}
\label{alg:algorithm2}
\textbf{Input:} \\
query $x$ \\
current community $c$\\
LLMs $\pi$, Knowledge Graph $\Omega$\\
maximum radius of subgraph $R_{max}$\\
number of candidate community $N$ 

\begin{algorithmic}[1] 
\STATE $g \leftarrow$ subgraph($c$, $\Omega$, $R_{max}$)
\STATE $C \leftarrow$ community\_detection($g$)
\STATE $C^{'} \leftarrow $ coarse\_pruning($C-H$, $c$) //based on modularity
\STATE $g_{text} \leftarrow$ Community2Text($g$)
\STATE $C^{''} \leftarrow $ fine\_pruning($C^{'}$, $x$, $\pi$, $N$)
\STATE return $C^{''}$
\end{algorithmic}
\end{algorithm}

\subsection{B. Details of Graph2Text}

\begin{figure}
  \includegraphics[width=\columnwidth]{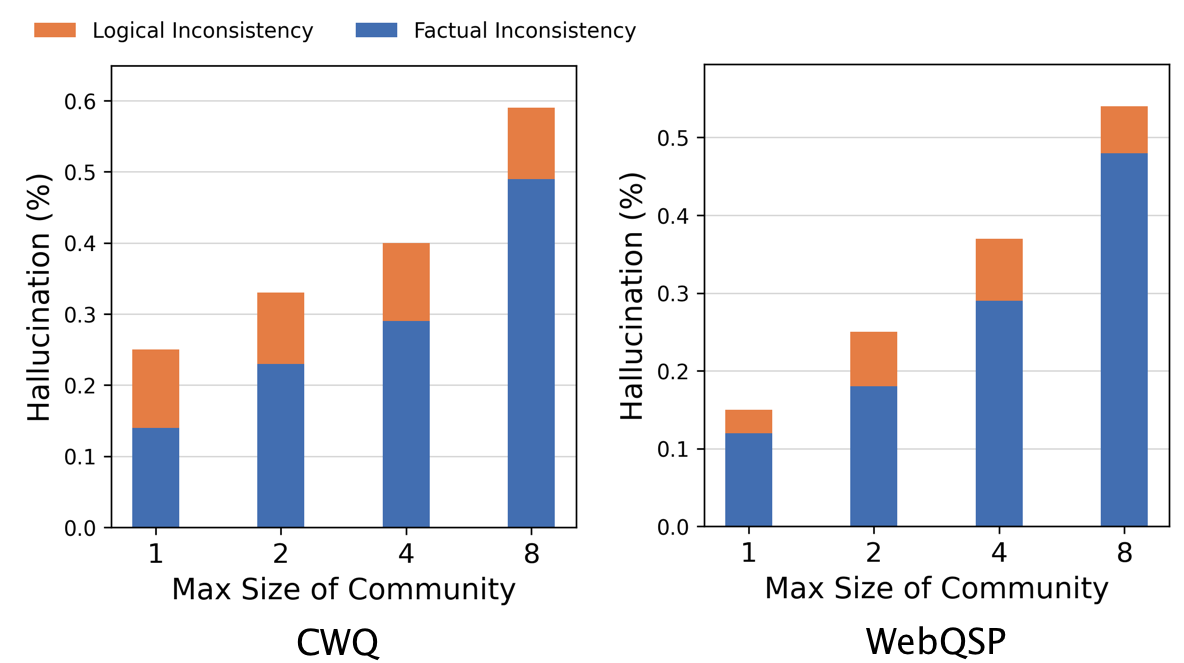}
  \caption{The hallucination ratio of Graph2Text Model on different Max Size of Community.}
  \label{fig:appendix_g2t_hallucination}
\end{figure}

\subsubsection{B1. Training Details}
We employed a straightforward fine-tuning approach using T5-base to implement Graph2Text. The fine-tuning data for Graph2Text comprised a combination of open-source datasets and AI generated datasets via prompt-based LLMs, with 10\% of the data allocated for validation. Statistics of fine-tuning and Examples of the data are listed in Tab \ref{tab:appendix_g2t_data_statistic} and Tab \ref{tab:appendix_g2t_data_example}, respectively.

For fine-tuning, apart from configuring the learning rate to 1e-5 and Epoch=2, all other parameters remained consistent with T5-base. We trained multiple models and ultimately selected the model based on Rouge1 as the filtering metric. The fine-tuning results are shown in Table \ref{tab:appendix_training_detail}. All G2T models were fine-tuned on an NVIDIA Tesla T4.

\begin{table}[]
\begin{tabular}{lllll}
\hline
Epoch & \begin{tabular}[c]{@{}l@{}}Training\\ Loss\end{tabular} & \begin{tabular}[c]{@{}l@{}}Validation\\ Loss\end{tabular} & Rouge 1 & \begin{tabular}[c]{@{}l@{}}Average\\ Length\end{tabular} \\ \hline
1     & 1.0038                                                  & 0.8319                                                    & 55.16   & 18.52                                                    \\
2     & 0.9221                                                  & 0.7903                                                    & 55.66   & 18.55                                                    \\ \hline
\end{tabular}
\caption{Training details of Graph2Text Model.}
\label{tab:appendix_training_detail}
\end{table}

\begin{table}[t!]
\small
\centering
\begin{tabular}{lllll}
\hline
                     &                                                                            & \multicolumn{3}{c}{Average}                                                                                                                                                                                                                       \\ \cline{3-5} 
                     & \multicolumn{1}{c}{\begin{tabular}[c]{@{}c@{}}Size\\ of Data\end{tabular}} & \multicolumn{1}{c}{\begin{tabular}[c]{@{}c@{}}Number\\ of Triples\end{tabular}} & \multicolumn{1}{c}{\begin{tabular}[c]{@{}c@{}}Tokens\\ of Triples\end{tabular}} & \multicolumn{1}{c}{\begin{tabular}[c]{@{}c@{}}Tokens\\ of Texts\end{tabular}} \\ \hline
\textbf{WebNLG}      & 15416                                                                      & 4.39                                                                            & 44.3                                                                            & 29.3                                                                          \\
\textbf{AI Labeling} & 12327                                                                      & 4.92                                                                            & 62.9                                                                            & 42.8                                                                          \\ \hline
\end{tabular}
\caption{Statistic of Data for fine-tuning Graph2Text Model.}
\label{tab:appendix_g2t_data_statistic}
\end{table}

\begin{table*}[!t]
\small
\centering
\begin{tabular}{ll}
\hline
\textbf{Triples}                                                                                                                                                                                                                      & \textbf{Texts}                                                                                                                                                                                                                                   \\ \hline
(Acharya Institute of Technology, nickname, ``AIT")                                                                                                                                                                                    & Acharya Institute of Technology is known as AIT.                                                                                                                                                                                                 \\ \hline
\begin{tabular}[c]{@{}l@{}}(United States, leaderName, Barack Obama),\\ (United States, capital, Washington, D.C.),\\ (1634: The Ram Rebellion, country, United States),\\ (United States, ethnicGroup, Asian Americans)\end{tabular} & \begin{tabular}[c]{@{}l@{}}1634 The Ram Rebellion was written in the U.S. \\ where the Native Americans are an ethnic group. \\ The capital of the country is Washington DC \\ and the President is Barack Obama.\end{tabular}                   \\ \hline
\begin{tabular}[c]{@{}l@{}}(A.E Dimitra Efxeinoupolis, location, Greece),\\ (Greece, currency, Euro),\\ (Greece, leader, Alexis Tsipras),\\ (Greece, leader, Prokopis Pavlopoulos),\\ (Greece, language, Greek language)\end{tabular} & \begin{tabular}[c]{@{}l@{}}Alexis Tsipras heads Greece where \\ Prokopis Pavlopoulos is also a leader. \\ The currency in Greece is the Euro \\ and Greek is spoken. \\ The A.E Dimitra Efxeinoupolis club \\ is located in Greece.\end{tabular} \\ \hline
\end{tabular}
\caption{Examplars of Data for fine-tuning Graph2Text Model.}
\label{tab:appendix_g2t_data_example}
\end{table*}

\begin{table*}[]
\small
\centering
\begin{tabular}{llll}
\hline
\textbf{Triples}                                                                                                                                                                                                                                                                                           & \textbf{Texts}                                                                                                                                                                                                                                                                                             & \textbf{Hallucination Type}                                     & \textbf{Reason}                                                                    \\ \hline
\begin{tabular}[c]{@{}l@{}}(The Gentlemen, cast member, \\ Charlie Hunnam),\\ (The Gentlemen, genre, action film),\\ (The Gentlemen, distributed by, Netflix),\\ (Go, genre, action film)\end{tabular}                                                                                                     & \begin{tabular}[c]{@{}l@{}}Charlie Hunnam is a cast member\\ of The Gentlemen. The Gentlemen is \\ an action film distributed by Netflix\\ and is also known as Go.\end{tabular}                                                                                                                           & \begin{tabular}[c]{@{}l@{}}Logical\\ Inconsistency\end{tabular} & \begin{tabular}[c]{@{}l@{}}LoGentlemen and Go\\ both are action film.\end{tabular} \\ \hline
\begin{tabular}[c]{@{}l@{}}(Charlie Hunnam, educated at, \\ University of Cumbria),\\ (Peter Strike, employer, University \\ of Cumbria),\\ (Portrait, Interior II* (triptych, centre panel), \\ location, University of Cumbria),\\ (Still Life Fruit, collection, \\ University of Cumbria)\end{tabular} & \begin{tabular}[c]{@{}l@{}}Charlie Hunnam was educated at the\\ University of Cumbria. Peter Strike is\\ employed by the University of Cumbria\\ and has a collection of Still Life Fruit. \\ The portrait ``Portrait, Interior II*" is \\ a triangular image of the University \\ of Cumbria.\end{tabular} & \begin{tabular}[c]{@{}l@{}}Factual\\ Inconsistency\end{tabular} & \begin{tabular}[c]{@{}l@{}}Portrait, Interior II* \\ is a location.\end{tabular}   \\ \hline
\end{tabular}
\caption{Exemplars of Hallucination Type of Graph2Text}
\label{tab:appendix_hallucination_type}
\end{table*}

\subsubsection{B2. Errors Analysis}
From the above experiments, we observe some cases where the T2T-based modes exceed the G2T-based ones. We hypothesize that the Hallucination generated by the Graph2Text model could have a negative impact on reasoning in LLMs, leading to a decrease in accuracy. To validate it, we manually analyzed 100 data samples drawn from each dataset and each max size of the communities for the statistic of hallucination. Fig. \ref{fig:appendix_g2t_hallucination} illustrates the percentage of the hallucinations across different max size of community. It can be observed that the proportion of hallucinations increases dramatically as the max size of community increases. This explains the existence of higher performance of t2t over g2t. After investigation, we generally categorize the hallucinations into two types: Factual Inconsistency and Logical Inconsistency \ref{tab:appendix_hallucination_type}. Factual Inconsistency refers to inaccuracies or fabrications in entity representations, while Logical Inconsistency pertains to errors in relationships between entities. Note that incorrect texts in our survey imply that hallucination is indeed present in the text, but not all the texts are hallucinatory, which explains the observation in Fig. 4 in the main paper that the accuracy does not drop dramatically even when the community is growing.

\subsection{C. Prompt Design}

\begin{figure}[!t]
  \includegraphics[width=\columnwidth]{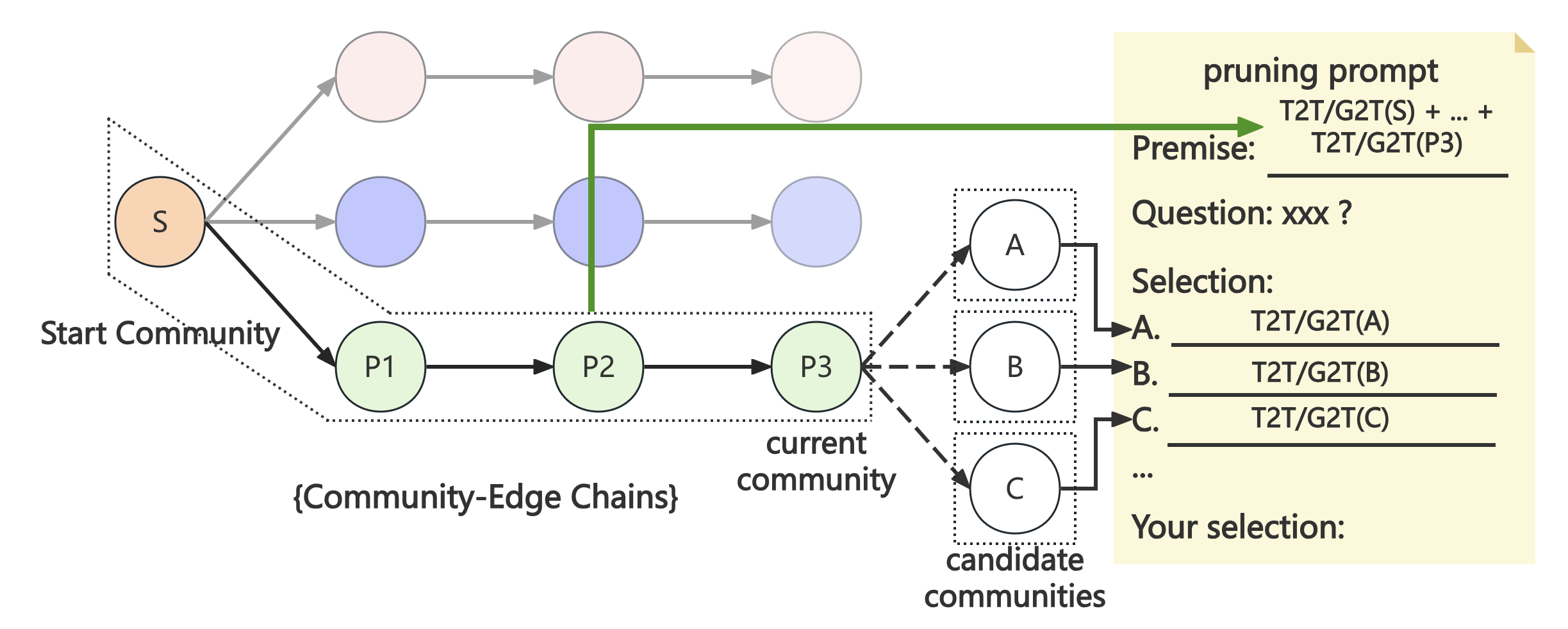}
  \caption{Diagram of construction of pruning prompt.}
  \label{fig:appendix_pruning_prompt}
\end{figure}

Unlike previous works, as FastToG regards community as the basic unit in the reasoning chain, prompts for the process of pruning and reasoning are redesign. All the prompts are open source and publicly available at xxx.

\subsubsection{C1. Pruning Prompt}
Pruning Prompt is designed to drive the LLMs to prune the communities with the potential to the answer. Fig. \ref{fig:appendix_pruning_prompt} shows the pruning process for 3 candidate communities based on the context communities S, P1, P2, and P3. To do this, ``Premise" in Fig. \ref{fig:appendix_pruning_prompt} is designed to carry the text which are converted from context communities. Similarly, candidate communities A, B, and C will be converted to texts which are placed behind ``Selection".
s
\subsubsection{C2. Reasoning Prompt}

\begin{figure}[t]
  \includegraphics[width=\columnwidth]{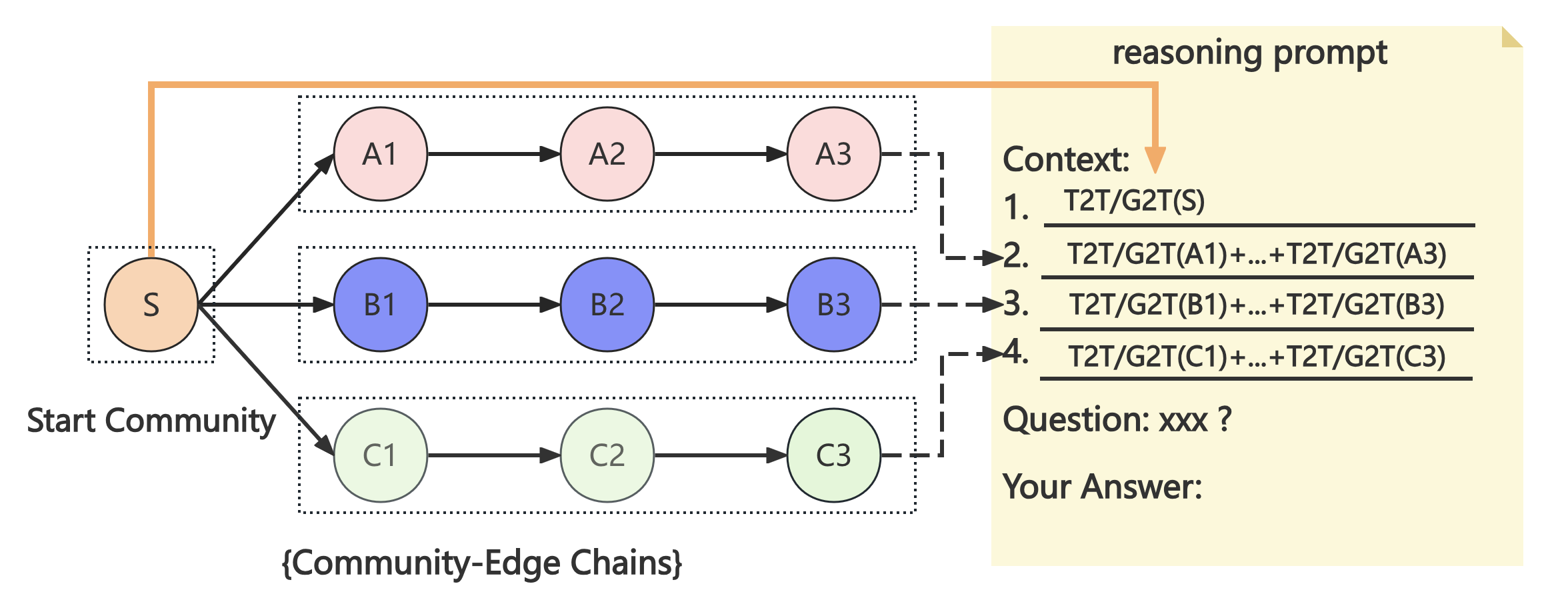}
  \caption{Diagram of construction of reasoning prompt.}
  \label{fig:appendix_reasoning_prompt}
\end{figure}

Reasoning Prompt is designed to prompt the LLMs to generate the answer based on the community chains started with ``Context". Fig. \ref{fig:appendix_reasoning_prompt} displays 3 community chains, each originating from community S. Adding community S to each community chain would inevitably leads to redundant information. Therefore, FastToG treats community S separately and inserts it as unique information at the beginning of the ``Context".

\end{document}